# 3D Human Action Recognition with Siamese-LSTM Based Deep Metric Learning

Seyma Yucer and Yusuf Sinan Akgul
VisLab, Department of Computer Engineering, Gebze Technical University, Kocaeli, Turkey
Email: {syucer, akgul}@gtu.edu.tr

*Abstract*—This paper proposes a new 3D Human Action Recognition system as a two-phase system: (1) Deep Metric Learning Module which learns a similarity metric between two 3D joint sequences using Siamese-LSTM networks; (2) A Multi-class Classification Module that uses the output of the first module to produce the final recognition output. This model has several advantages: the first module is trained with a larger set of data because it uses many combinations of sequence pairs. Our deep metric learning module can also be trained independently of the datasets, which makes our system modular and generalizable. We tested the proposed system on standard and newly introduced datasets that showed us that initial results are promising. We will continue developing this system by adding more sophisticated LSTM blocks and by cross-training between different datasets.

## I. INTRODUCTION

Human action recognition is one of the most popular topics in computer vision and machine learning. The analysis of 2D and 3D video sequences [1] enables many real-life applications, such as entertainment and multimedia, surveillance, healthcare, robotics and so on [2], [3]. In the last decade, lots of advanced solutions on 2D video datasets have been proposed [4], [5].

As RGB+D video capturing devices become more ubiquitous and cheaper, action recognition studies focus on 3D action data [6]. Depth information is the effective way of representing the structure of real-world scenes and objects [7]. It is especially very effective in recovering the 3D human skeletal joint positions using common RGB+D devices such as MS Kinect. 3D joint information is crucial for the task of human action recognition because humans perform their actions using their joints [8]. For example, walking action involves foot and knee joint movements, eating action involves hand joint movements in 3D space. Therefore, joints may represent human actions better [9] hence human action recognition using 3D joint position data gets increasingly more attention from the researchers.

Skeleton-based action recognition studies can be classified into two categories, hand-crafted feature-based methods and deep network-based methods. Unlike hand-crafted feature-based methods [11]–[13], deep network-based methods [12]–[16] can make a human action classification using the features learned directly from the data.

For the recent years, deep learning-based methods achieved outstanding results. These methods generally take the 3D skeleton sequence data of an action as the input and produce an action class label as the output. While some deep learning methods consider the temporal input as static [17], the others use temporally sensitive deep learning methods such as LSTM and RNN [14], [16].

However, all of them consider this problem as a classification task from the raw 3D skeleton frames. This end-to-end classification approach creates some problems for the deep learning systems because these systems need huge amounts of data for the training which is usually very difficult to obtain for the 3D skeleton data.

In this paper, we pose the 3D human action recognition task as a Deep Metric Learning (DML) [18] problem which learns a similarity metric between two 3D joint sequence data using deep learning methods. One can compare two different 3D joint sequences using the automatically learned metric which can later be used for the classification of the compared sequences. The main advantage of this approach is that; we argue that it is easier to learn a similarity metric on smaller datasets than learning a classifier because it is possible to train the DML network with many different combinations of the available sequences.

To the best of our knowledge, this is the first work that uses DML for the 3D human action recognition task. There are methods that use manually designed similarity metrics for the action recognition tasks [19], but deep learning based metric learning techniques was not tried. This is surprising given that DMLs are commonly used for the biometric identification problems [20] and re-identification problems.

Our DML network employs a Siamese-LSTM (S-LSTM) structure, which repeats the same network two times in parallel with the shared parameters. Each one of our networks employs the same LSTM architecture which is very popular for the human action recognition tasks because LSTMs can learn temporal sequence information efficiently.

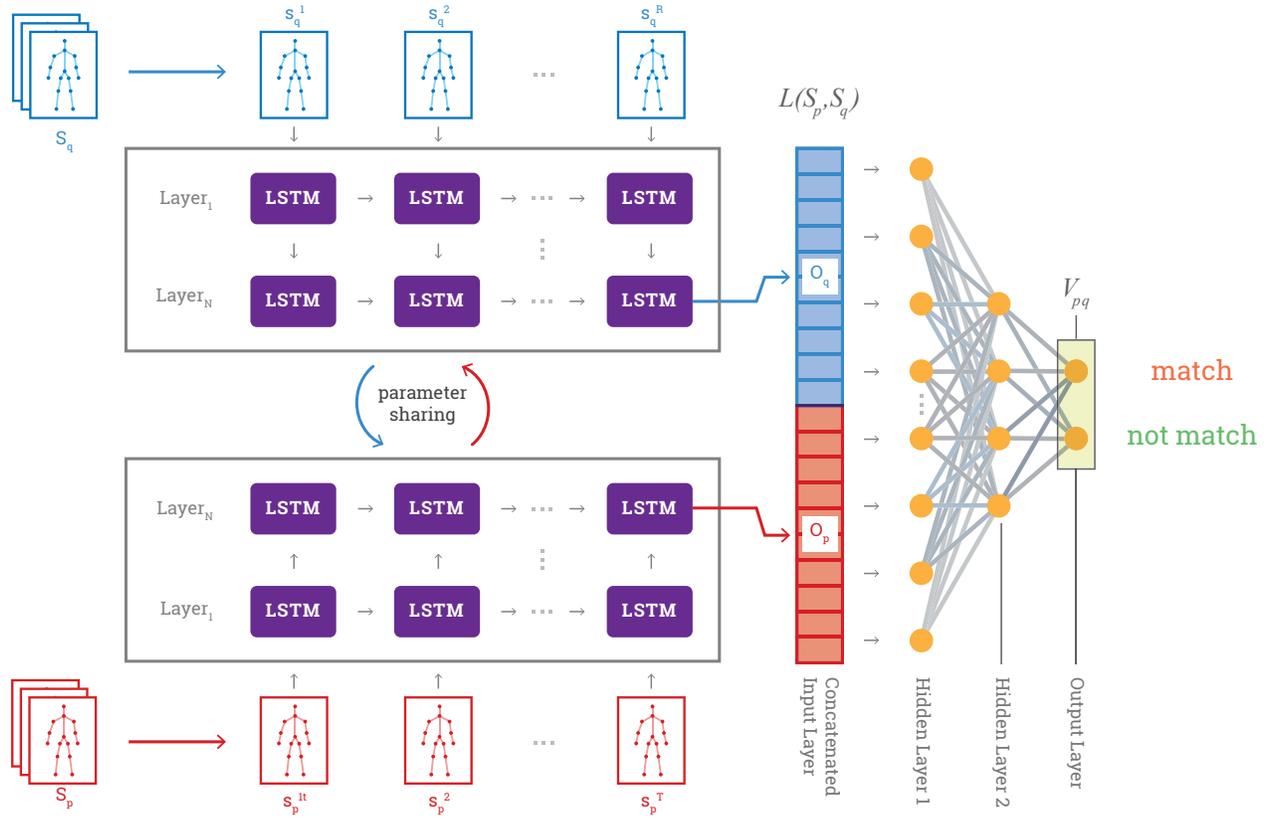

Figure 1. S-LSTM-based Deep Metric Learning Module

Our proposed classification system employs two successive modules. The first module, Siamese-LSTM is for the similarity metric calculation between action pairs. The second module, multi-class classification (MCC) is for the real action classification.

Our DML approach is not restricted by the initial action classes in the training set because we are not using classification for the fixed number of classes in S-LSTM module. Instead, this module learns similarities between action sequences. Therefore, our S-LSTM module can be trained with many action pairs from different datasets, which makes our method more generalizable because learning the similarity across many different datasets is supposed to perform better than learning this information from a single set. As expected, S-LSTM module of our system has a considerable impact on the recognition accuracy. Note that our system can employ any type of LSTM or RNN based networks inside the Siamese-LSTM DML module such as [21]. This makes our proposed system more generalizable and modular.

This paper also introduces GTUAction3D dataset to validate our proposed method on a new smaller sized training set even though our method can also be used on large-scale datasets.

The rest of the paper is organized as follows. Section 2 describes the proposed method in three parts; (1) Siamese LSTM based DML module, (2) Multi-class classification module and (3) Module training. Section 3 explains our setup and our experimental results on given datasets. The last section concludes our work as well as the future works based on our system.

II. PROPOSED METHOD

Our method contains two modules; Siamese LSTM (S-LSTM) module and Multi-class Classification (MCC) module.

A. *Siamese LSTM-based DML Module*

For the 3D human action recognition task, the final goal is to find the action class given 3D skeleton frame sequences as accurately as possible. However, we claim that learning a similarity metric between the action sequences offers many advantages. Therefore, we propose an S-LSTM network to take 3D action pairs as inputs and learn a similarity metric between them.

Fig. 1 shows the general structure of the proposed metric learning module. This module takes two 3D action sequences $S_p = \{s_p^1, s_p^2, s_p^3, ... s_p^T\}$ and $S_q = \{s_q^1, s_q^2, s_q^3, ... s_q^R\}$, which are ordered. $T$ and $R$ are the number of frames for each sequence.

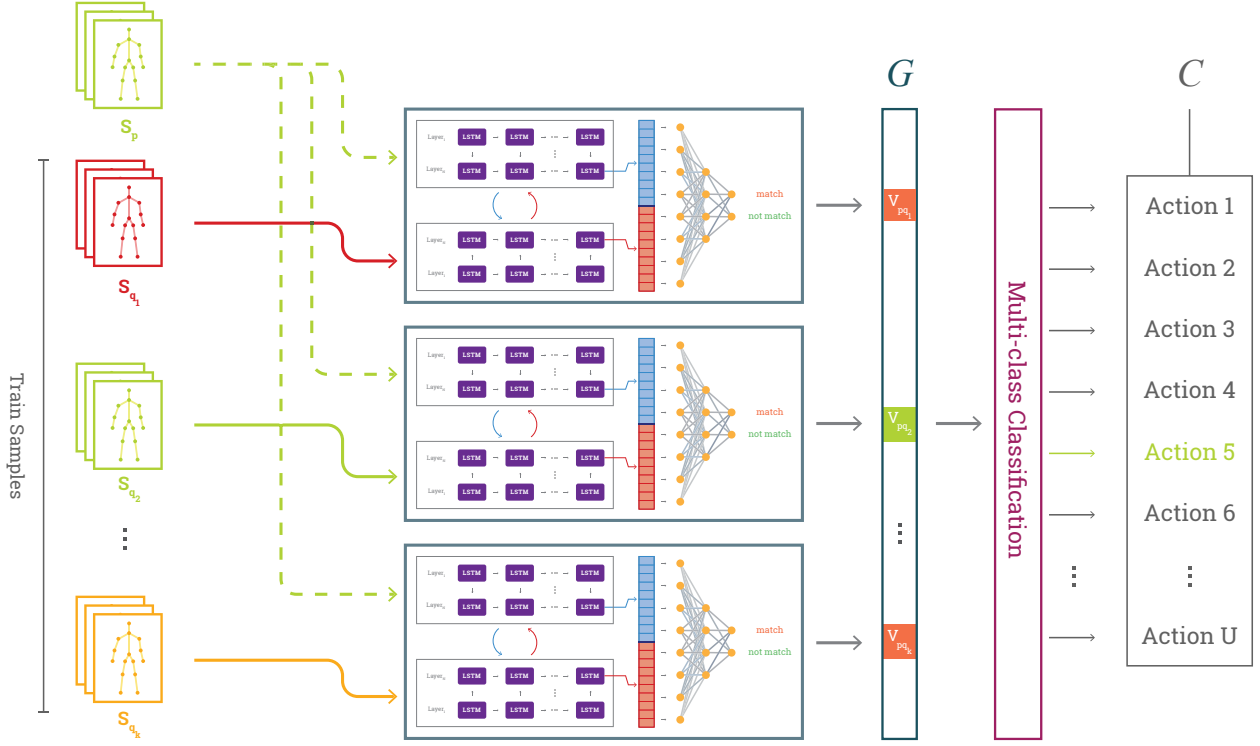

Figure 2. Multi-class classification module

$s_p^t = \{j_1^t, j_2^t, ... j_N^t\}_p$ is a single skeleton frame where $N$ is the total number of 3D joints in a single frame at time t. $j_n^t = \{x_n, y_n, z_n\} \in R^3$ is the coordinate for single joint $j_n^t$. Note that $T$ and $R$ are different for each action sequences. Therefore, we use LSTM cells as the basic building block of our metric learning system. Each of the two LSTM networks takes one sequence as input and they produce two output vectors $O_p \in R^M$ and $O_q \in R^M$. Note that sizes of these vectors are fixed regardless of the number of frames ($T$ and $R$) in the input sequences.

We model the block of LSTM modules with a function $L(S_p, S_q) \in R^{2M}$ that returns a vector which is concatenations of the vectors $O_p$ and $O_q$. $L(S_p, S_q)$ holds extracted deep similarity features of the input sequences. We feed this vector to a Multi-Layer Perceptron (MLP) to produce one hot vector $V \in R^2$ that assigns on of the (match, not match) labels.

$$D(L(S_p, S_q)) = V_{pq}, \quad (1)$$

where $D$ is the model for the MLP, which in our case has two hidden layers.

$$V_{pq} = Softmax(b^3 + W^3 Re(b^2 + W^2 Re(b^1 + W^1 L(S_p, S_q)))), (2)$$

where $W$'s are the network weighs, $b$'s are the bias terms, $Re$ (ReLU) is the rectified linear activation unit. The effectiveness of the module $D$ is important as it highly influences the accuracy of the multiclass classification module. This will be explained in the next section.

### B. Multi-class Classification Module

As described before, the final output of the 3D action recognition systems should be an action class label. The output vector of S-LSTM model is a 2-dimensional match-no match vector, which is not sufficient for this label assignment.

The task of the MCC module in Fig. 2 is to get results of comparison between a test action sequence $S_p$ and many other train sequences $S_{q_1}, S_{q_2}, S_{q_3}, ... S_{q_k}$ where $k$ is the number of such training sequences. Let $G \in R^{2k}$ be the concatenation of the vectors from the S-LSTM model results $V_{pq_1}, V_{pq_2}, V_{pq_3}, ... V_{pq_k}$. The vector $G$ is fed to the MCC as input and the output of this module is a one hot vector of $C \in R^u$, where u is the number of action classes to be recognized. We tested different methods for the MCC such as KNN and SVM.

Note that although the previous module S-LSTM needs to be trained with action pairs as input and the match-no match labels as output, our MCC module needs to be trained with the final action class labels. Therefore, S-LSTM can be trained with action sequence pairs from different datasets but MCC module needs to be trained on a single dataset with its own action class labels.

*C. Module Training*

Training of S-LSTM and MCC is done separately. For the training of S-LSTM, there is a label imbalance problem because the number possible no match pairs is much larger than the matching pairs. To keep the training set of the SLSTM in balance, for a dataset of U action classes, we keep the ratio of match/no match pairs around 1/U.

Our MCC module does not have any label imbalance problem because we expect that the number action labels to be relatively uniforms across the action classes. Note however that the number training samples for the MCC module is much less than the number of training samples for the S-LSTM, which is one of the main contributions of our method.

III. EXPERIMENTS

We test our proposed method in two datasets: Florence Action 3D dataset [22] and our new GTU-Action dataset. We also compared the performances of the SLSTM and MCC modules to understand their interaction better.

We have implemented the proposed system using Tensor Flow and performed all trainings and experiments on a NVidia GTX 1080 GPU board with 8GB of memory.

*A. S-LSTM and MCC Modules Accuracy*

As mentioned before, our system can recognize human actions as well as can learn the deep metrics between them, so that it can find the similarity between actions.

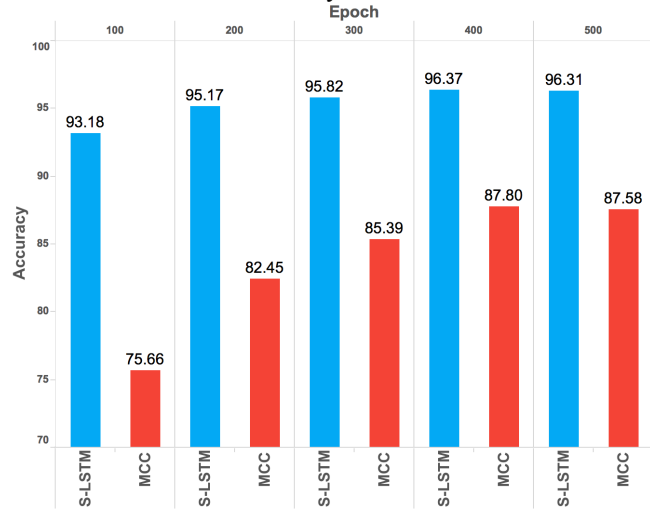

Figure 3. Comparison between Siamese-LSTM Module Accuracy and Recognition Accuracy on Florence Action 3D Dataset

To amplify the significance of the metric learning accuracy over recognition accuracy, we validated on both test sets and the results are given below in Fig. 3 and Fig. 4 As expected, the performance of both S-LSTM and MCC modules increases as the number of epochs increase up to some epoch number.

We also observe that small increases in the performance of the SLSTM module is reflected by large increase in the MCC performance. This means that a properly trained S-LSTM with robust LSTM blocks should help our final recognition performance significantly. Therefore, for the future work, we plan to try state of the art LSTM blocks and use training data from multiple datasets.

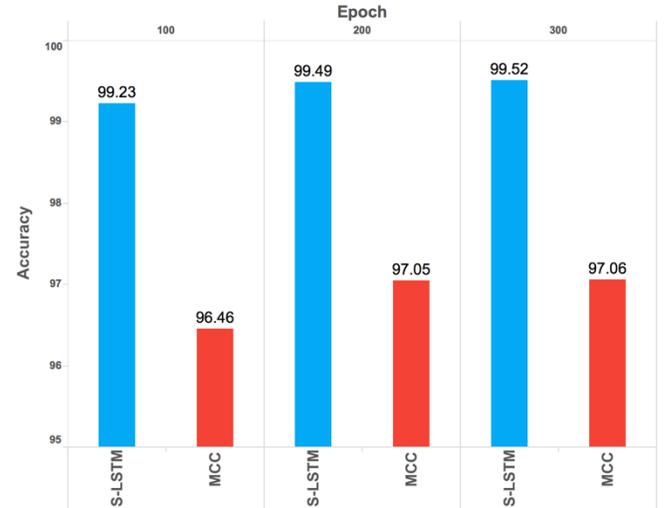

Figure 4. Comparison between Siamese-LSTM Module Accuracy and Recognition Accuracy on GTU Action Dataset

*B. Florence 3D Action Dataset*

We evaluate the performance of our method on the Florence 3D Action dataset [22]. Our method achieves 89.51% accuracy on cross-subject tests.

We also applied standard classification methods as given in Table I which shows that our proposed method significantly increases the base LSTM method performance. We expect that, using more sophisticated LSTM methods, such as [13] as the base method in our S-LSTM, we can achieve much better results. Even with this configuration, we are comparable with many state of the art methods.

Since Florence 3D Action dataset has limited number of actions with short frame lengths, a simpler model with smaller number of parameters and layer count works better. Therefore, we use 1-layer LSTM block inside our model.

In this dataset, the model architecture is as follows: note that the stream consists of two branches until concatenation.

{LSTM (75,128,2) - CONCAT (256,1) - FC (256,128)- ReLU- FC (128, 64)- ReLU-FC (64,2)}

Using early stopping, we stopped training process in 400th epoch for subject 2,3,7. For the rest of subjects, we trained our model 500 epochs.

TABLE I. RESULTS ON THE FLORENCE ACTION 3D DATASET

| Methods | Accuracy |
|---|---|
| Multi-part Bag-of-Poses [22] | 82.00% |
| Riemannian Manifold [23] | 87.04% |
| Latent Variables [24] | 89.67% |
| Lie Group [9] | 90.88% |
| Feature Combinations [25] | 94.39% |
| SVM | 23.30% |
| Softmax | 61.61% |
| 1-Layer LSTM | 76.99% |
| 2-Layer LSTM | 72.32% |
| **Siamese-LSTM DML** | **89.51%** |

## C. GTU Action 3D Action Dataset

Our dataset consists of 508 action samples from 14 different type of actions from 9 subjects. These actions are (1) arm closing, (2) right arm closing, (3) waist stretch, (4) walking, (5) right leg bending, (6) left leg bending, (7) left arm closing, (8) right-left step, (9) crouching, (10) sit up, (11) sit and applause, (12) lower back, (13) right and left 8-Step, (14) neck relaxation. These actions are mostly indoor sports actions recorded with a MS Kinect II sensor.

On average, our method classified 97.06% of actions correctly. As can be seen from Table II, our method exceeded standard methods by a significant margin.

The confusion matrix in Fig. 5 explains our method's behaviour on this dataset. According to the confusion matrix, walking action (4th action) is confused most with one right-left step action (8th action).

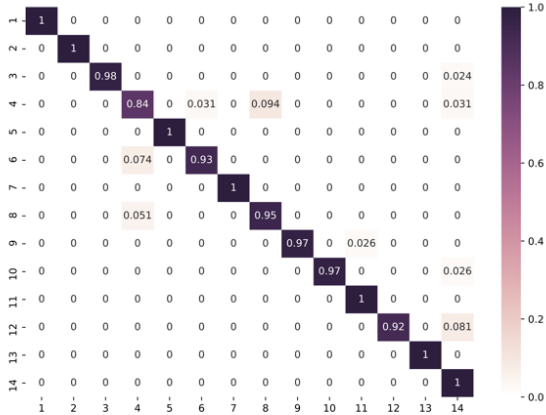

Figure 5. Confusion Matrix of GTU Action 3D Dataset with Siamese-LSTM DML method

In this dataset, the model architecture is as follows:

{LSTM (75,200,2) - LSTM (200,200,2) - CONCAT (400,1) - FC (400,300)-ReLU- FC (300, 150)- ReLU-FC (300,50) - ReLU-FC (50,2)}

For the all of subjects, we trained our model 300 epochs.

TABLE II. RESULTS ON THE GTU ACTION 3D DATASET

| Standard Methods | Accuracy |
|---|---|
| SVM | 48.04% |
| SOFTMAX | 75.99% |
| 1-Layer LSTM | 90.21% |
| 2-Layer LSTM | 95.47% |
| **Siamese-LSTM DML** | **97.06%** |

## IV. CONCLUSION

We introduced a new 3D Human Action Recognition system that uses a deep metric learning module as the main engine. This module is trained using pairs of action sequences that makes the training data set much larger, which is critical because it is difficult to obtain training data for 3D action recognition systems. Our metric learning system does not have to be trained on a single dataset, which makes our system more generalizable and portable between different applications. The experiments performed on standard and novel datasets showed that the initial results are comparable with the state of the art recognition systems. For future work, we will focus on employing more advanced LSTM blocks in our Siamese networks. We also plan to demonstrate the effectiveness of our system in training on more than one dataset.

**Seyma Yucer** received the BS degree in computer engineering from Gebze Technical University, Kocaeli, Turkey in 2015. She is currently working as a research assistant in the Institute of Information Technologies, Gebze Technical University, Turkey. She is also continuing to work towards her Ms degree in the Department of Computer Engineering at Gebze Technical University, Turkey. Her research interests include application of computer vision in human action recognition, 3D video and sequential data analysis.

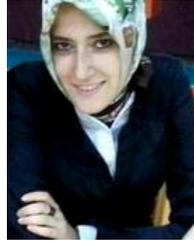

**Yusuf S. Akgul** received the BS degree in computer engineering from Middle East Technical University, Ankara, Turkey in 1992. He received the MS and PhD degrees in computer science from University of Delaware in 1995 and 2000, respectively. He worked for Cognex Corporation, Natick, MA between 2000 and 2005 as a senior vision engineer. He is currently with the Department of Computer Engineering, Gebze Technical University, Turkey. His research interests include application of computer vision in medical image analysis, video analysis, object recognition, 3D analysis, and industrial inspection.

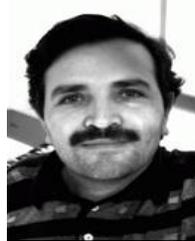